\pdfoutput=1

\documentclass[11pt]{article}

\usepackage[preprint]{acl}

\usepackage{times}
\usepackage{latexsym}

\usepackage[T1]{fontenc}

\usepackage[utf8]{inputenc}

\usepackage{microtype}

\usepackage{inconsolata}
\usepackage{paralist}

\usepackage{graphicx}
\usepackage{multirow}
\usepackage{booktabs}
\usepackage{xspace}
\usepackage{array}
\newcolumntype{P}[1]{>{\centering\arraybackslash}p{#1}}

\title{The Unreasonable Effectiveness of Model Merging for Cross-Lingual Transfer in LLMs}

\author{\textbf{Lucas Bandarkar}\thanks{Correspondence: \href{mailto:lucasbandarkar@cs.ucla.edu}{lucasbandarkar@cs.ucla.edu}}
    \ \ \ \ \ \ \ \ \ \ \ \hfill
    \textbf{Nanyun Peng} \\
    University of California, Los Angeles \\
}

\newcommand{\llama}{\textsc{Llama} 3.1\xspace}
\newcommand{\falcon}{\textsc{Falcon} 3\xspace}
\newcommand{\qwen}{\textsc{Qwen}2.5\xspace}
\newcommand{\aya}{\textsc{Aya} Expanse\xspace}
\newcommand{\belebele}{\textsc{Belebele}\xspace}
\newcommand{\flores}{\textsc{FLoRes}\xspace}

\newcommand{\ls}{Layer-Swapping\xspace}

\begin{document}
\maketitle

\begin{abstract}

Large language models (LLMs) still struggle across tasks outside of high-resource languages. In this work, we investigate cross-lingual transfer to lower-resource languages where task-specific post-training data is scarce.
Building on prior work, we first validate that the subsets of model parameters that matter most for mathematical reasoning and multilingual capabilities are distinctly non-overlapping.
To exploit this implicit separability between task and target language parameterization, we develop and analyze numerous \textit{modular} frameworks to improve the \textit{composition} of the two during fine-tuning. These methods generally employ freezing parameters or post hoc model merging to assign math and language improvement to different key parts of the LLM. In the absence of in-language math data, we demonstrate that the modular approaches successfully improve upon baselines across three languages, four models, and two fine-tuning paradigms (full and LoRA). Furthermore, we identify the most consistently successful modular method to be fine-tuning separate language and math experts and model merging via \ls \citep{bandarkar2025layer}, somewhat surprisingly.
We offer possible explanations for this result via recent works on the linearity of task vectors. We further explain this by empirically showing that reverting less useful fine-tuning updates after training often outperforms freezing them from the start.

\end{abstract}

\section{Introduction}

Post-training large language models (LLMs) on labeled text data is a critical step in developing models for real-world applications.
However, when these LLMs are fine-tuned for lower-resource languages, significant challenges arise due to the pretrained model's limited capabilities.
Although in recent years the broader scaling of pretraining and increased investment in additional languages \cite{ayaexpanse, llama3} have led to major improvements, pretrained LLMs still struggle to understand and generate text in all but a few languages \citep{romanou2025include, multilingualsurvey}.

This pretraining disparity is further exacerbated by the lack of available high-quality multilingual fine-tuning data \citep{singh-etal-2024-aya} and the significant cost to procure such annotated data (even through machine translation). For many of the capabilities developers target during post-training (e.g., instruction-following, reasoning, or safety) there are only sufficient open-source data available in English, Chinese, and a handful of other languages. This motivates the need for better cross-lingual transfer: the generalization of learned capacities from high-resource languages to lower-resource ones \citep{pmlr-v119-hu20b, philippy-etal-2023-towards}. 

Despite recent releases of massive mixture-of-expert LLMs \citep{ mosaic2024dbrx, deepseekai2025deepseekv3technicalreport, qwen3}, a large majority of modern LLMs are \textit{dense}, meaning that all parameters are active during training and inference. However, even within dense LLMs, recent works have found separability in where and how varying capabilities are represented \citep{yin2024lofit, yunzhiknowledge}. For example, multilingual capabilities are typically concentrated in the top and bottom transformer layers and multi-head attention parameters of an LLM \citep{chang-etal-2022-geometry, choenni-etal-2024-examining}. This notably contrasts mathematical reasoning capabilities being encoded mainly in the middle transformer layers  \citep{hanna2023how, stolfo-etal-2023-mechanistic}. In the context of cross-lingual transfer, this functional separation motivates \textit{modular} approaches to fine-tuning, which distinct model components can be trained, swapped, or merged \citep{Bengio2020A, pfeiffer2023modular} for efficient and flexible multi-objective optimization.


\begin{figure}[t]
    \centering
    \includegraphics[width=8cm]{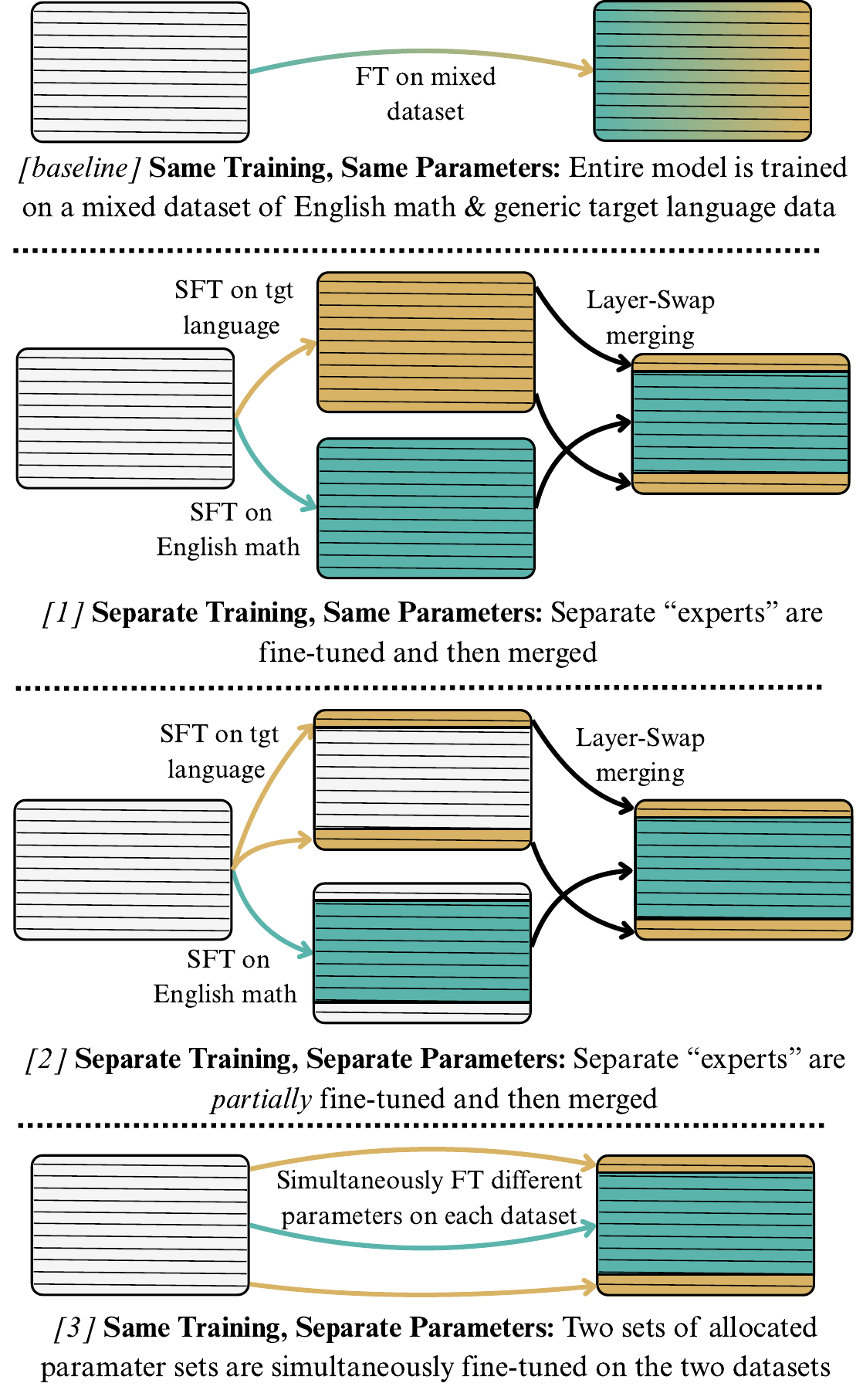}
    \caption{Illustration of the three methods that induce modularity by imposing target language capabilities (brown) and mathematical reasoning (blue) on separate LLM parameters. \textit{[1]} is from \citet{bandarkar2025layer}}
    \label{teaser}
\end{figure}

In this work, we explore several modular approaches for composing target task and target language capabilities in off-the-shelf dense LLMs.
Our goal is to induce modularity by exploiting the differences in parameters that are most relevant to mathematical reasoning versus multilingual capabilities. We focus on the prevalent scenario where task-specific data is scarce in the target language but readily available in English. We address this by working with two datasets; one English math dataset for supervised fine-tuning (SFT) and one general, multi-task SFT dataset in the target language. Using the target languages of Bengali, Swahili, and Telugu, we evaluate the methods on the multilingual math benchmark, MGSM \citep{shi2023language}. 

With these datasets, we evaluate numerous training paradigms that incentivize the model, to varying degrees, to learn multilingual or math capabilities in specific parameters. We organize the settings along two axes: (1) whether the models are optimized separately or together over the two SFT datasets and (2) whether the same or separate model parameters are trained on the datasets. When the models are trained separately, we combine the learned capabilities using model merging methods such as variants of \ls \citep{bandarkar2025layer}. To train separate model parameters, we start by dividing all parameters into two partitions according to prior work: (1) one set allocated to target language training and (2) one set to English math. Only allocated parameters are fine-tuned, while the opposite partition is frozen. We additionally develop a method to train separate parameters in a single, joint training by frequently freezing and unfreezing parameters to simulate simultaneous training.

Despite the strong starting capabilities of the four LLMs and the data-constrained setting, our experimental results show that all of the modular solutions outperform our baselines, despite being subject to varying training constraints. This implies that intentional separation of parameters and/or training improves the \textit{compositionality} of task and language capabilities. 

Amongst our modular solutions, we surprisingly find that post hoc model merging via \ls outperforms more coordinated multi-task fine-tuning approaches. To contextualize this counterintuitive result, we explore recent academic literature that help explain the phenomenon.  We provide empirical evidence for training all model parameters, even if large portions will be discarded during \ls. While these subsets of task vectors are unproductive, freezing them during fine-tuning leads to less optimal updates to the target parameters. Notably, we rationalize that the fine-tuning task vectors ($\Delta$s) are quite linear within individual parameter blocks \citep{dai2025leveraging},  meaning they can be added, scaled, or interpolated as linear components \citep{adilova2024layerwise}.

Overall, we enumerate the following principal contributions of this work:
\begin{compactitem}
    \item We develop and synthesize a number of modular solutions that \textit{each} increase compositionality for cross-lingual transfer compared to non-modular baselines, demonstrated through extensive experiments.
    \item Of the modular methods, we find that fine-tuning all parameters and then merging via \ls performs best on average.
    \item We provide a mix of theoretical and empirical explanations to explain the surprising success of \ls relative to alternatives.
\end{compactitem}

\begin{table*}[ht]
    \begin{center}
    \begin{tabular}{|p{3.5cm}||c|P{2.3cm}P{2cm}|P{1.5cm}P{2cm}|}
        \toprule
        \textbf{Training Description} & \textbf{Base Model}& \textbf{Partial LoRA} & \textbf{Partial SFT} &  \textbf{LoRA} & \textbf{Full SFT} \\
        \midrule
        \textbf{Math-only}        & 19.0\% & 18.0\% & 19.5\% & 18.9\% & 19.6\% \\
        \textbf{Language-only}    & 19.0\% & 19.2\% & 19.8\% & 19.7\% & 20.3\% \\
        \textbf{Data mixing}      & 19.0\% & -      & -      & 19.7\% & 20.4\% \\
        \textbf{Simultaneous SFT} & 19.0\% & 20.4\% & 21.0\% & -      & -      \\
        \textbf{\ls}              & 19.0\% & 20.0\% & 20.4\% & 20.8\% & \textbf{21.5\%} \\
        \bottomrule
    \end{tabular}
    \caption{Summary Table of Results. Each value represents \textit{the average across four models, three languages, and multiple training runs} on MGSM in 2-shot evaluations. The last row represents ``Separate Training'' while the ``Partial'' trainings correspond to ``Separate Parameters'' trainings. All results shown here and in all other tables of this paper display exact-match (EM) accuracy ($\uparrow$) as a percentage.}
    \label{summary}
    \end{center}
\end{table*}

\section{Background}

\subsection{Cross-Lingual Transfer}

The relative abundance of textual data available in English in comparison to other languages has long motivated research in developing methods to efficiently transfer learned capabilities across languages \citep{koehn-knight-2002-learning}. 
Typically, some capabilities transfer naturally across languages, as evidenced by the superior performance of multilingual models on low-resource languages compared to monolingual models \citep{firat-etal-2016-multi, pires-etal-2019-multilingual, artetxe-etal-2020-cross}. 
In encoder models, the text embedding could be aligned across languages to improve transfer using methods such as contrastive learning
\citep{mikolovembeddings, artetxe2018unsupervised, muller-etal-2021-first}.  

However, cross-lingual alignment in more modern decoder-only models has become less methodical because of the lack of universal embedding \citep{kargaran2024mexamultilingualevaluationenglishcentric}. Since most popular LLMs have been trained on a majority English corpora, recent works have examined how much intrinsic cross-lingual transfer occurs at different training stages \citep{choenni-etal-2023-languages, wang-etal-2024-probing-emergence}. These large models have broader generalization and robustness, but still fail to transfer much of their capabilities across languages \citep{philippy-etal-2023-towards}. Recent works have identified prompting methods \citep{shi2023language, zhang-etal-2024-plug} or post-training data augmentation \citep{dang-etal-2024-rlhf, she-etal-2024-mapo, lai-etal-2024-llms} to help generalization.

\subsection{Modularity in Multilingual NLP} \label{multilmodularity}
A major constraint for models being able to process many languages has been the number of parameters available to represent them. As a result, improving a language model in one language risks undermining its knowledge of another, termed the \emph{curse of multilinguality} \citep{conneau-etal-2020-unsupervised, pfeiffer-etal-2022-lifting}. Naturally, numerous methods have been proposed to increase the model's parametric capacity without increasing the inference cost, such as mixture-of-expert architectures \citep{fedus-switch} that route tokens according to their language \citep{nllb-2022-no}.
Methods that leverage modular parameters were developed to compose capabilities for transfer learning by inserting trainable adapters within model layers \citep{houlsby19a, pfeiffer-etal-2021-adapterfusion}. These methods were modified for multilinguality by allocating adapters for particular languages and switching them in or out depending on the input \citep{bapna-firat-2019-simple, pfeiffer-etal-2020-mad}. \citet{pfeiffer-etal-2022-lifting} extended these methods by pretraining an adapter-based multilingual model from scratch. In decoder models, cross-lingual adapters have also been proposed at the token embedding level \citep{jiang2025frankenadapter}.

Even in dense LLMs, however, interpretability research has identified the emergence of effective modularity \citep{csordas2021are} 
as LLM parameters scale \citep{zhang-etal-2022-moefication, qiu-etal-2024-unlocking, chen2025emergenceabstract}. 
Principally, numerous recent works have identified that just a few transformer layers at the top and bottom of English-centric LLMs are responsible for multilingual capabilities, notably by mapping input and output into a universal representation \citep{kojima-etal-2024-multilingual, wendler-etal-2024-llamas, tang-etal-2024-language, alabi-etal-2024-hidden, wu2025the}. Similar patterns are observed in modern sparse mixture-of-experts LLMs, where it is also observed that language-specialized experts are completely distinct from task/domain-specialized ones \citep{bandarkar2025multilingualmoe}.

\subsection{Model Merging}
Model merging is the practice of combining the weights of multiple checkpoints of the same model architecture into a singular model. While averaging models is a fundamental machine learning approach to increase statistical robustness \citep{breiman96}, the averaging of model checkpoints, dubbed a model soup by \citet{Wortsman2022}, has re-emerged in large-scale LLMs as a method to increase model robustness. More importantly, it also increases the search space of valid model variants at any given training step without additional costly training runs \citep{llama3}. However, simple weight averaging is vulnerable to negative transfer, or interference,
between checkpoints so numerous methods have been presented to selectively merge parameters \citep{Ilharco2023, Yadav2023, yu2024language}. Surprisingly, training models on separate data and then merging can often outperform a single training run on mixed data \citep{tang2024merging, aakanksha2024mixdatamergemodels} and has shown to be highly effective in large-scale multilingual pretraining \citep{ayaexpanse}. For cross-lingual transfer in particular, \citet{ansell-etal-2022-composable} showed that sparse fine-tuning can lead to better composition. \citet{bandarkar2025layer} extended this by notably identifying that mathematical reasoning was concentrated in parameters different from multilingual capabilities. As a result, model variants trained on English math data and multilingual data can be combined by \ls, or swapping the transformer layers most important to each.

\section{Experimental Setup} \label{setup}

\subsection{Evaluation} Limited by the lack of task-specific benchmarks for medium- and low-resource languages, we focus on MGSM \citep{shi2023language} as the target task of this project. MGSM is a mathematical reasoning benchmark parallel across 10 languages as a result of high quality translations from the popular English benchmark, GSM8K \citep{cobbe2021gsm8k}. For MGSM, we report exact match accuracy in two-shot, as one- and zero-shot led to inconsistent results. More few-shot examples did not display substantial gain. For target languages, we choose the languages in MGSM where the four LLMs perform the worst: Bengali, Telugu, and Swahili. In addition, the lack of open-source math SFT data available in these languages motivates the need for more effective cross-lingual transfer. For a given fine-tuned model, we also evaluate the two-shot MGSM performance in English to evaluate its math performance irrespective of target language capability. Conversely, we use the multilingual MCQA benchmarks \textsc{Global} MMLU \citep{singh2025globalmmlu} and \belebele \citep{bandarkar-etal-2024-belebele} as pure language understanding signals, independent of math.

\subsection{Models} We run experiments on four state-of-the-art instruction-finetuned LLMs: \falcon 7B \citep{Falcon3}, \qwen 7B Instruct \citep{qwen2.5}, \llama 8B Instruct \citep{llama3}, and \aya 8B \citep{ayaexpanse}. All have similarly high performance on MGSM in English. \llama and \falcon are English-centric, \qwen bilingual with Chinese, and \aya explicitly multilingual. However, all officially cover numerous other languages (up to 23 for \textsc{aya}) and perform reasonably on such languages, which we verify using \belebele and \textsc{Global} MMLU. Bengali, Swahili, and Telugu are amongst the official languages for none of these models. As a result, the four models are all low-scoring in MGSM in these languages, with the exception of \textsc{Llama} on Swahili (See Appendix~\ref{shelf}).

\subsection{Parameter Allocation} \label{paramalloc}
To determine which parameters to ``allocate'' to each capability, we rely on a mix of interpretability papers and small-scale empirical tests. As mentioned in Section~\ref{multilmodularity}, numerous papers have identified the most important parameters for multilingual capabilities to be the first few and last few transformer layers of LLMs. These works, however, typically discuss mostly English-centric models (such as \llama and \falcon). We therefore need to evaluate this for bilingual and multilingual models like \qwen and \aya. For mathematical reasoning, we note that \citet{bandarkar2025layer} identifies the middle and late-middle transformer layers as being the most important. This work, and numerous others 
\citep{voita-etal-2019-analyzing, ma-etal-2021-contributions, zhao2024how}, similarly identifies multi-head attention parameters as critical to multilingual capabilities, as opposed to multi-layer perceptron parameters.

\begin{table}[ht!]
    \begin{center}
    \begin{tabular}{|p{4.2cm}||P{1.2cm}|P{1.1cm}|}
        \toprule
        \textbf{Parameters that are frozen or reset} & \textbf{Frozen during} & \textbf{Reset \hskip 1cm after} \\
        \hline
        base (no SFT)                         & 78.4\% & 78.4\% \\
        \midrule
        \textit{[Z]} only top-4 and bottom-8 layers (inverse of intuition) & 78.2\% & 78.9\% \\
        \midrule
        \textit{[A]} all MHA parameters + MLP parameters in top-2 and bottom-6 layers & 79.4\% & 79.8\% \\
        \hline
        \textit{[B]} only top-4 and \hskip 1cm bottom-8 layers & 79.8\% & 79.8\% \\
        \hline
        \textit{[C]} only top-2 and \hskip 1cm bottom-6 layers & 79.7\% & 80.0\% \\
        \midrule
        None                                   & 80.1\% & 80.1\% \\

        \bottomrule
    \end{tabular}
    \caption{MGSM 2-shot results ($\uparrow$) on the \textit{English} split after SFT on the English math data averaged across four models. These results (1) validate that our intuition leading to our parameter allocations \textit{[A, B, C]} is reasonable seeing as results are close to full fine-tuning and are significantly higher than the inverse allocation \textit{[Z]}. Additionally, (2) these results demonstrate that full fine-tuning then reverting parameters (second column) is more effective than freezing those parameters from the start (first column).}
    \label{mathfrozen}
    \end{center}

\end{table}

To empirically verify these assumptions on our selected models, we run SFT over our datasets with different subsets frozen. We evaluated numerous ways to partition the parameters and find a number of splits that enable improvements on English math and on language-specific signals (e.g. \belebele).
To validate that the good performance when freezing parameters is because the trainable parameters are particularly useful for a target task, we also run experiments with the \textit{opposite} allocation (e.g. middle layers frozen during mathematical reasoning training) and find that it works poorly. 

While the search space of which parameters to freeze is large, we settle on three partitions that show sufficient empirical success:

\begin{compactitem}
    \item \textit{[A]} All multi-head attention parameters allocated to the target language. Then, amongst the multi-layer perceptron parameters, those in the first six and last two transformer layers still allocated to language, while those in the rest of the 32- or 36-layer LLM for math.
    \item \textit{[B]} The first eight and last four transformer layers allocated to language, the rest for math.
    \item \textit{[C]} The first six and last two transformer layers allocated to language, the rest for math.
\end{compactitem}

In these three settings, both mathematical reasoning and target language capabilities improve similarly to full SFT with a fraction of trainable parameters (See Table~\ref{mathfrozen} for results for math). We evaluate the three for each of our experimental settings and, unless noted, report the highest scoring.

\subsection{Training} For SFT data, we create four datasets, one for math in English and one instruction dataset for each of the three target languages. The math instruction dataset consists of English math word problems from the Orca-Math synthetic dataset \citep{mitra2024orcamath}. For the language datasets, we replicate the creation of ``generic'' instruction fine-tuning datasets from \citet{bandarkar2025layer} by combining samples from open-source instruction and task-specific datasets. Importantly, there are no math samples in these multi-task language datasets. We provide specific details and citations for these data collections in Appendix~\ref{sftdata}.

Due to constraints on the amount of verifiable-quality data available in each of the target languages, our datasets are controlled at 80k samples, 2k of which is reserved for validation. Because of significantly diminishing returns exhibited by the validation loss and downstream evaluations, we only train for one epoch for each of our settings.

We additionally duplicate all experiments using Low-Rank Adapters (LoRA) \citep{hu2022lora}. Specifically, we use rank-stabilized LoRA \citep{kalajdzievski2023rslora} applied to both multi-layer perceptron and multi-head attention parameters. In general, the adjustments of our methods to be compatible with LoRA were minor unless noted otherwise. With four models, three languages, and two fine-tuning approaches (full and LoRA), we have a total of 24 experimental settings. For each, we do hyperparameter search over several runs to ensure comparability (See Appendix~\ref{reproduc} for details).

\section{Experiments} \label{settings}

We describe numerous methods that modularize off-the-shelf, dense LLMs in different ways. We describe \textit{separate training} as when we conduct separate SFT runs on different datasets, albeit starting from the same off-the-shelf model. As previously mentioned, the separately trained checkpoints are then merged via \ls. \textit{Separate parameters} implies that only the partition of parameters \textit{allocated} (See Section~\ref{paramalloc}) to that dataset are trained while the rest remain frozen.

\subsection{Baselines (Math-only and Language-only)}
\label{base}

For comparison, we evaluate a number of straightforward SFT setups to serve as baselines. We do full-parameter training runs for each of the target language generic SFT datasets and the English math SFT dataset. For further baselines, we re-run the above when leaving only parameters \textit{allocated} to that capability trainable, and the rest are frozen. In addition, we replicate both full training and partial training in LoRA, where parameters are ``frozen'' if no adapter is added for that parameter.
\subsection{Data Mixing (Same Training, Same Parameters)}

As an additional baseline, we randomly mix the two datasets together and jointly optimize over the two disjoint tasks with all parameters left trainable.

\subsection{\ls (Separate Training, Same Parameters)} \label{separsame}

For this setting, we exactly recreate the method presented by \citet{bandarkar2025layer}. Starting from the same base model, separate variants are trained on different tasks, dubbed ``experts''. Concretely, one expert has been trained on the English math data, and the other on the target language instruction dataset. To recompose a single model, the top and bottom transformer layers from the target language expert replace those in the math expert, while the math experts' middle layers remain. We additionally implement the equivalent of this methodology with LoRA, where the set of adapters is merged by combining the adapters corresponding to parameters that would be swapped. Note that we do not retrain these experts and simply use the checkpoints from our baseline trainings.

\subsection{\ls with Partial SFT (Separate Training, Separate Parameters)} \label{separsepar}

We modify \ls so that only the parameters involved in the model merging are trained, and all those eventually ignored are kept frozen during training. The idea for this is that no parameters are unnecessarily trained and we can incentivize the training to focus the learned capabilities into the desired parameters. Similar to above, we do not retrain experts and simply merge checkpoints from our frozen parameter baselines.

\subsection{Simultaneous Partition SFT (Same Training, Separate Parameters)} \label{samesepar}
We design a methodology to ``simultaneously" fine-tune two partitions of LLM parameters on two different datasets. To do so, we apply a gradient step on a batch from one dataset on the corresponding partition of parameters. Then, we switch which parameters are frozen and sample a batch from the other dataset for the next gradient step. This frequent back-and-forth is intended to ensure the coordination of parameter updates during multi-task optimization. The validation set contains an equal amount from each datasets.

\paragraph{Switching} We default to a single step before switching to best simulate fully simultaneous training, but additionally experiment with more steps between. We set the effective batch size
\footnote{Effective batch size is the product of the batch size per GPU, number of GPUs, and gradient accumulation steps.}
to 64. At the end of each step, all parameters just updated are frozen for the next step and conversely, all frozen parameters are unfrozen. In addition, a flag for the data iterator is switched to ensure the next batch of data will be sampled from the appropriate dataset. For LoRA training, the same logic is implemented.

\paragraph{Optimizer} We consider numerous approaches to adapt the AdamW optimizer \citep{loshchilov2018decoupled} used in all previous experiments. Although we technically employ a single optimizer initialized on all parameters during training, we configure it to function as two independent optimizers, each exclusively managing its own separate subset of parameters. Namely, when a subset of parameters $A$ is frozen, the corresponding AdamW optimizer states $\Omega_A$ (momentum and variance estimates) are also frozen in time. As a result, when the parameters in $A$ are unfrozen, the corresponding momentum and variance estimates of $\Omega_A$ still reflect only the gradients steps previously applied to $A$. However, the other parameters $A^c$ have been updated in the meantime, meaning $\Omega_A$ risks being outdated given the modified loss landscape. To test the impact of this inconsistency, we ablate over different numbers of steps between switches and find that the differences are very negligible (See Appendix \ref{ablations}). We conclude that the optimizer restarting on an outdated loss landscape is of minimal concern, presumably because of the smoothness of the loss topology. Since there is a single optimizer, the learning rate schedule is the same for all (constant with warmup). 
And while the gradients tend to be larger for the multilingual data, we set a maximum gradient norm of $1.0$ for clipping. 

\begin{table*}[ht]
    \begin{center}
    \begin{tabular}{|l||c|c||c|c|c|c|c|c|}
        \toprule
        \multicolumn{9}{c}{Performance Comparison of Modular Solutions} \\
        \midrule
        \multirow{2}{*}{\textbf{SFT Type}} & \multirow{2}{*}{\textbf{Base}} & \multirow{2}{*}{\textbf{Full}} & \multicolumn{2}{c|}{\textbf{Simultaneous SFT}} & \multicolumn{4}{c|}{\textbf{\ls}} \\
        &  & & \small \textbf{Full} & \small \textbf{LoRA} & \small \textbf{Full SFT} & \small \textbf{LoRA} & \small \textbf{Part. SFT} & \small \textbf{Part. LoRA} \\
        \midrule
        \textbf{Swahili}        & 23.5\% & 25.1\% & 25.9\% & 25.2\% & \textbf{26.7\%} & 25.8\% & 25.1\% & 24.8\% \\
        \textbf{Bengali}        & 25.6\% & 27.9\% & 27.9\% & 26.9\% & \textbf{28.7\%} & 27.5\% & 27.0\% & 26.7\% \\
        \textbf{Telugu}         & 7.9\%  & 8.2\%  & \textbf{9.3\%}  & 9.0\%  & 9.2\%  & 9.2\%  & 9.0\%  & 8.6\%  \\
        \midrule
        \textbf{English}        & 78.4\% & 80.4\% & 81.8\% & 80.5\% & 80.9\% & 80.8\% & 79.9\% & 80.0\% \\
        \textbf{sw,bn,te AVG}   & 19.0\% & 20.4\% & 21.0\% & 20.4\% & \textbf{21.5\%} & 20.8\% & 20.4\% & 20.0\%\\
        \bottomrule
    \end{tabular}
    \caption{All values presented above are MGSM 2-shot EM accuracy ($\uparrow$), averaged across four models. The baseline presented for comparison in the 3rd column is the full SFT on the mix of the two datasets.}
    \label{modularresults}
    \end{center}
\end{table*}

\section{Results} \label{results}

Our experimental setting was designed to replicate a real-world scenario where multilingual LLM developers would take a post-trained LLM and are limited by the amount of in-language post-training data. This constrained scenario means only modest improvements are achievable.
However, we do observe several conclusive patterns. Across our different four models and three languages (12 \textit{conditions}), we can summarize into 6 \textit{treatments} discussed in Sections~\ref{base} to~\ref{samesepar}. Despite the small magnitude of differences, the rank-based Friedman test (non-parameteric) shows statistically significant differences between the \textit{treatments} at the 0.05 significance level. 

In our setting, we find that only training on the language dataset is more effective in improving the target language MGSM score than only on the math dataset (details in Appendix~\ref{baselines}). This implies, perhaps, that what our four models need most, is improved Swahili, Bengali, or Telugu abilities as opposed to math improvement.

We validate the lack of need for full-parameter training when doing both language adaptation and math SFT. Once the most useful parameters have been identified for such a skill, as discussed in Section~\ref{paramalloc}, comparable performance to full SFT can be achieved with a fraction of the trainable parameters. Beyond potentially contributing to compositionality, this leads to faster and more memory-efficient training. More details on these baselines can be seen in Appendix~\ref{baselines}. We do note, however, that in the absence of resource limitations, SFT with less trainable parameters converged a bit slower and full fine-tuning still performed best. This is also true for LoRA, which has much less trainable parameters by nature.

A significant result is that all our modular solutions perform statistically-significantly better than the non-modular baselines, as can be seen in Table~\ref{summary}. This is strongly the case for Telugu and Swahili in the displayed four-model averages, but varies more by specific modular method for Bengali in comparison to the top baseline (data mixing) (See Appendix~\ref{bargraph} for per-language results).

Within our modular solutions, however, we find numerous surprising results. First, freezing the unused parameters in training experts before \ls does not improve upon full training. As detailed in the last four columns of Table~\ref{modularresults}, the difference in performance is better when all modules are being finetuned for both LoRA and full-parameter SFT (statistically significant). This is counter-intuitive because the layers eventually merged are potentially dependent on parameter changes that are being replaced. Second, \ls surprisingly outperforms the simultaneous SFT. This is surprising because in our simultaneous SFT, the modularity is being imposed cohesively as opposed to the ad hoc merging of layers from separate training runs. We note, however, that the simultaneous SFT performs second-best.

To validate results further, we also evaluate more expensive Continual Pretraining (CPT) for \qwen in Bengali across the experimental designs and find agreement with our SFT results (See details in Appendix~\ref{cptresults},~\ref{cptdata}). However, we limit discussion of these results because of the small scale of experimental results.

We additionally analyze the composability of individual experts under \ls. We define a good merging indicator as an evaluation signal \textit{of an expert} that correlates with the performance of the merged model. We find that performance on general NLU benchmarks—\belebele and \textsc{Global} MMLU—is a stronger indicator of a \textit{language} expert’s merge quality than MGSM results in the target language. Similarly, MGSM performance in English is a better predictor for a \textit{math} expert than MGSM in the target language.
This is notable because MGSM in the target language is the target task of course, yet results more directly related to the training data tends to be more important for proper task composition. 
\section{Discussion} \label{discussion}

Given the rejection of our hypothesis that simultaneous fine-tuning would most effectively compose task and language capabilities, we discuss potential explanations for this outcome.

\paragraph{Train-then-Revert vs. Freeze-then-Train} Intuition may dictate that fine-tuning parameters and then later reverting part of them should be less effective than simply freezing those parameters from the start. In the former, the fine-tuning is unaware of future edits while the latter provides hard constraints during optimization. However, empirically, we find that across models, training-then-resetting outperforms freezing-then-resetting. We display this for our English math fine-tuning in Table~\ref{paramalloc}. This explains why \ls with full training (Section~\ref{separsame}) may be preferential to solutions involving freezing parameters. We conclude that while a large portion of fine-tuning weight updates are not needed in the end, either because they are noisy or redundant \citep{supermario}, they enable optimization in a very high-dimensional space. This is analogous to recent papers discussing the Lottery Ticket Hypothesis \citep{frankle2018the}, where it has been concluded that training a full neural network and then pruning it leads to stronger models than the same pruning before training \citep{frankle2021pruning}.

\paragraph{Concatenating Components in \ls} We seek to explain why concatenating transformer layers from separately fine-tuned ``experts" is so seamless.
Task vectors \citep{ilharco2023editing} are the $\Delta$s that result from fine-tuning (i.e., $\theta_{FT} - \theta_0$). Task vector
\textit{linearity} refers to the property that linear combinations of such task vectors form a coherent, effective model. \citet{ortizjimenez} identifies that linearized task vectors exhibit better mergeability. Meanwhile, when fine-tuning heavily post-trained models like those used in our experiments, recent works show that updates to individual model layers exhibit significant linearity \citep{zhou2024on, razzhigaev-etal-2024-transformer, dai2025leveraging}. 
Furthermore, research on mode connectivity \citep{pmlr-v119-frankle20a, NEURIPS2018_be3087e7} shows individual transformer layers can be smoothly interpolated \citep{zhou2023going, adilova2024layerwise}. These works provide explanation for why ad hoc \ls is not more degradative.

\paragraph{Further Considerations} We note that model merging is convenient because the configuration (e.g., what parameters to swap), can be determined after training. This enables fast iteration through configurations without retraining. This flexibility is sacrificed for our ``separate parameters" methods, which require fixing parameter allocations. However, an inconvenience of merging methods is the need to train two experts, potentially doubling the amount of training runs for hyperparameter search.

\section{Conclusions} \label{conclusions}

Our results demonstrate that imposing modularity into dense LLMs for cross-lingual transfer is quite effective in low-data scenarios. We empirically validate this with numerous ways to impose such modularity through fine-tuning with frozen parameters or model merging, all of which prove more effective than non-modular baselines. Furthermore, we discover the surprising success of \ls over other modular methods that fine-tune task and language together or do not ad hoc revert parameter updates. We conjecture that the success of this ad hoc merging method is because the math and language experts, when represented as task vectors, exhibit a high degree of linearity. As a result, this method benefits from more robust training over all parameters while also leading to effective compositionality. We also empirically demonstrate that the success of \ls is in part due to frozen-parameter fine-tunings underperforming full fine-tunings followed by parameter resets.
\section{Future Work} \label{futurework}




We encourage further work in multilingual NLP that leverages implicit modularity in LLMs, induces it during training, or designs explicitly modular architectures. Our parameter allocation strategy relied on previous interpretability work and limited empirical evidence, and the search space of modular configurations is largely unexplored. With post hoc model merging, iterating through many ablations can be quick. Although we focused on mathematical reasoning—due to limited multilingual task-specific datasets—future work should examine other tasks that may warrant different parameter allocations. More broadly, these results underscore the importance of improving interpretability around how capabilities are parameterized in LLMs, such as multilinguality. If we can better localize and separate parameters by function, our findings suggest that modularization may yield significant improvements.

\section*{Limitations}

\paragraph{Small $\Delta$s} Our decision to use the instruction fine-tuned version of each of the open-source LLMs for our experiments was a conscious one that came with many considerations. We prioritized replicating a  real-life practical scneario, where model developers would start from already fine-tuned LLM versions because of their broader capabilities. However, as a result, this meant that our fine-tuning experiments only led to relatively small performance improvements with respect to the starting checkpoint. Such checkpoints have undergone extensive post-training, notably with significant mathematical reasoning samples and varying amounts of multilingual samples. Therefore, possible model improvements with these small datasets were small, risking results that were not statistically significant. Nevertheless, this allowed us to control for the amount of improvement on benchmarks that was simply a result of the LLMs' improved ability to follow instructions after SFT, in addition to reflecting a more practical setting.

\section*{Acknowledgement}
The authors acknowledge the support provided by Tanmay Parekh and Mohsen Fayyaz for this project.

\bibliography{anthology,custom}

\captionsetup[table]{skip=12pt}
\captionsetup[figure]{skip=8pt}
\setlength{\belowcaptionskip}{5pt}
\onecolumn
\appendix
\section{Appendix}

\subsection{Detailed Baseline Results} \label{baselines}
\begin{table}[htbp]
    \begin{tabular}{|l||c||c|c|c|c|c|c|c|c|c|}
    \toprule
    \multicolumn{10}{c}{Detailed Performance of Non-Modular Baselines} \\
    \midrule
    \textbf{SFT Dataset} & \textbf{None} & \multicolumn{2}{c|}{\textbf{Data-Mixing}} & \multicolumn{3}{c|}{\textbf{Math-Only}} & \multicolumn{3}{c|}{\textbf{Language-Only}}  \\
    \textbf{SFT Type} & \small \textbf{Base} & \small \textbf{Full} & \small \textbf{LoRA} & \small \textbf{Full} & \small \textbf{LoRA} & \small \textbf{Part. FT} & \small \textbf{Full} & \small \textbf{LoRA} & \small \textbf{Part. FT}\\
    \midrule
    \textbf{Swahili}        & 23.5\% & 25.1\% & 24.8\% & \textbf{25.2\%} & 24.4\% & 25.0\% & 24.8\% & 23.8\% & 24.3\% \\
    \textbf{Bengali}        & 25.6\% & 27.9\% & 26.0\% & 26.1\% & 24.8\% & 25.6\% & \textbf{28.3\%} & 26.6\% & 26.9\% \\
    \textbf{Telugu}         & 7.9\%  & 8.2\%  & 8.4\%  & 7.4\%  & 7.4\%  & 8.0\%  & 7.9\%  & \textbf{8.6\%}  & 8.2\%  \\
    \midrule
    \textbf{English}        & 78.4\% & 80.4\% & 80.0\% & \textbf{81.3\%} & 81.0\% & 80.6\% & 79.9\% & 78.8\% & 79.0\% \\
    \textbf{sw,bn,te AVG}   & 19.0\% & \textbf{20.4\%} & 19.7\% & 19.6\% & 18.9\% & 19.5\% & 20.3\% & 19.7\% & 19.8\% \\
    \bottomrule
    \end{tabular}
    \caption{All values presented above are MGSM 2-shot EM accuracy ($\uparrow$), averaged across four models. Generally, we find that data mixing is the most effective, but with very small difference in comparison to language-only SFT. We exclude Partial LoRA results for space considerations, but report here that the results were for all numbers, 0-1\% lower than LoRA results.}
\end{table}

\subsection{CPT Results for \qwen in Bengali} \label{cptresults}
\begin{table}[htbp]
    \begin{tabular}{|l||c||c|c|c|c|c|c|c|c|c|c|}
    \toprule
    \multicolumn{10}{c}{Detailed Performance of CPT Experiments} \\
    \midrule
    \textbf{SFT Dataset} & \textbf{None} & \textbf{Mix} & \multicolumn{2}{c|}{\textbf{Math-Only}} & \multicolumn{2}{c|}{\textbf{Lang-Only}} & \textbf{Simult.} & \multicolumn{2}{c|}{\textbf{\ls}}  \\
    \textbf{SFT Type} & \small \textbf{Base} & \small \textbf{Full}  & \small \textbf{Full} & \small \textbf{Part.FT} & \small \textbf{Full} & \small \textbf{Part.FT} & \small \textbf{Part.FT} & \small \textbf{Full} & \small \textbf{Part.FT}\\
    \midrule
    \textbf{Bengali} & 37.6\% & 38.2\% & 33.2\% & 34.2\% & 37.6\% & 37.8\% & 38.8\% & 39.4\% & 38.8\% \\
    \midrule
    \textbf{English} & 76.8\% & 77.6\% & 80.0\% & 79.8\% & 74.0\% & 73.8\% & 80.2\% & 79.2\% & 79.6\% \\
    \bottomrule
    \end{tabular}
    \caption{All values presented above are MGSM 2-shot EM accuracy ($\uparrow$), averaged across two runs. We find that our main results from SFT mostly stand, but limit our conclusions as the small number of runs prevent the findings from being statistically significant. We note that CPT trainings more substantially degrade performance in the \textit{opposite} capability than SFT. ``Mix" is ``Data-Mixing" and ``Simult." is ``Simultaneous FT", shortened for space.}
\end{table}

\subsection{ Number of Gradient Steps Between Switches} \label{ablations}

\begin{table}[h!]
    \begin{center}
    \caption{Ablation over the number of gradient steps to do on a single dataset and single partition of model parameters before switching back to the other data and parameters. All runs were controlled to have the same exact hyperparameter settings on \qwen 7B Instruct with the target language Swahili. Four upper layers and eight lower layers were allocated for the target language, and a learning rate $1.2e^{-06}$}
    \begin{tabular}{|P{3.5cm}||P{3.5cm}|P{3.5cm}|P{3.5cm}|}
    \hline
    \textbf{Gradient Steps per Switch} & \textbf{Starting Validation Loss} & \textbf{Ending Validation Loss} & \textbf{$\Delta$ for MGSM, Swahili} \\ \hline
    $1$     & $2.301$ & $1.605$ & $+3.2\%$ \\
    $5$     & $2.301$ & $1.612$ & $+2.4\%$ \\
    $10$    & $2.301$ & $1.613$ & $+2.8\%$ \\
    $50$    & $2.301$ & $1.613$ & $+2.0\%$ \\
    $200$   & $2.301$ & $1.602$ & $+0.8\%$ \\
    $500$   & $2.301$ & $1.565$ & $+1.2\%$ \\
    $1171$  & $2.301$ & $1.536$ & $-1.2\%$ \\

    \hline
    \end{tabular}
    \end{center}
\end{table}

These results indicate no negligible differences between the tested step counts. This implies the concern discussed in Section~\ref{samesepar} of the optimizer unfreezing with an outdated loss landscape is minimal. Or at least, it implies that the ability to do numerous steps without interruption in the same setting outweighs this concern.
And while increasing the gradient steps per switch does provide no negligible difference on the validation loss, intuitively it leads to a training paradigm farther from a truly simultaneous training. We find that on the target task, MGSM in Swahili, performance goes down progressively as the gradient steps per switch is increased. This implies the composition of math and Swahili capabilities are working less effectively.

\subsection{Details for Reproducibility} \label{reproduc}

For reproducibility, we detail our implementation and hyperparameters for training. The datasets themselves are outlined in Sections~\ref{sftdata}~and~\ref{cptdata}. 

\begin{itemize} 
    \item Training is run on a single cluster of A100s, typically with only one GPU per training run.
    \item Training methods are developed using the trl python package \citep{vonwerra2022trl} and models accessed via HuggingFace.
    \item Learning rate ranged across training runs, but was typically in the range $\left[1.0, 2.0\right] \times 10^{-6}$. 
    \item For LoRA, it ranged from $\left[4.0, 9.0\right] \times 10^{-6}$. Rank and Alpha parameters were either $(64,16)$ or $(32,8)$.
    \item Sequence length was either 512 or 1024. Effective batch size was typically 32, except for effective batch size of 64 for simultaneous training, as described in Section~\ref{samesepar}.
    \item Evaluation is performed using the Language Model Evaluation Harness \citep{eval-harness}.
\end{itemize}

\subsection{Bar Graph of Per-Language Results} \label{bargraph}

\begin{figure}[h!]
    \includegraphics[width=15cm]{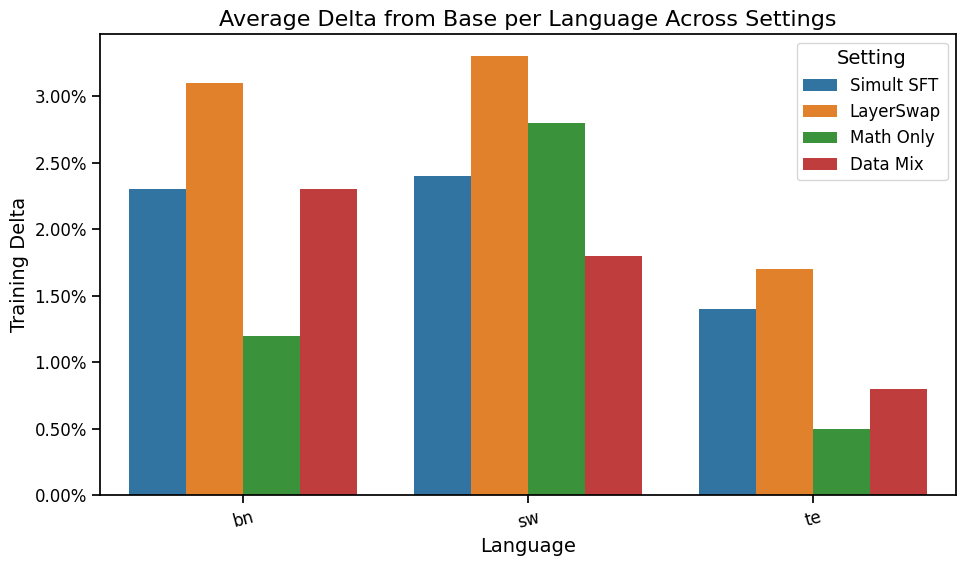}
    \caption{Per-language breakdown of the average performance gain seen during our different types of training, averaged across four models. We see that while math-only SFT (green) does well for Swahili and mixed-data SFT (red) does well for Bengali, our two modular solutions work consistently well across the three languages. Note: the y-axis is a percentage because the evaluation score is accuracy, \textit{not} because this table displays percent change.}
\end{figure}

\vspace{1cm}

\subsection{SFT Datasets} \label{sftdata}
\begin{table}[h!]
    \begin{center}
    \caption{Datasets used for supervised-fine-tuning (SFT) in this project}
    \begin{tabular}{|c|p{6cm}|p{7cm}|}
    \hline
    \textbf{Category} & \textbf{Datasets} & \textbf{URL} \\ \hline
    
    \multirow{1}{*}{Math} 
     & Orca Math word problems dataset from Microsoft \citep{mitra2024orcamath} & \footnotesize \url{https://huggingface.co/datasets/microsoft/orca-math-word-problems-200k} \\ \hline
    
    \multirow{3}{*}{Telugu} 
     & Aya Dataset from Cohere for AI \citep{singh-etal-2024-aya} & \footnotesize \url{https://huggingface.co/datasets/CohereForAI/aya_dataset} \\ \cline{2-3}
     & NLLB English-Telugu translation data from FAIR \citep{nllb-2022-no} & \footnotesize \url{https://huggingface.co/datasets/allenai/nllb} \\ \cline{2-3}
     & Synthetic English instruction dataset, machine translated to Telugu by Telugu-LLM-Labs & \footnotesize \url{https://huggingface.co/collections/Telugu-LLM-Labs/indic-alpaca-datasets-65f2a3687d5cdbce8880c581} \\ \hline 
    
    \multirow{4}{*}{Bengali} 
     & Aya Dataset by Cohere for AI \citep{singh-etal-2024-aya} & \footnotesize \url{https://huggingface.co/datasets/CohereForAI/aya_dataset} \\ \cline{2-3}
     & NLLB English-Bengali translation data from FAIR \citep{nllb-2022-no} & \footnotesize \url{https://huggingface.co/datasets/allenai/nllb} \\ \cline{2-3}
     & IndicShareLlama dataset from AI4Bharat \citep{khan-etal-2024-indicllmsuite} & \footnotesize \url{https://huggingface.co/datasets/ai4bharat/indic-align} \\ \cline{2-3}
     & BongChat dataset from Lumatic AI & \footnotesize \url{https://huggingface.co/datasets/lumatic-ai/BongChat-v1-253k} \\ \hline
    
    \multirow{4}{*}{Swahili} 
     & Aya Dataset by Cohere for AI \citep{singh-etal-2024-aya} & \footnotesize \url{https://huggingface.co/datasets/CohereForAI/aya_dataset} \\ \cline{2-3}
     & NLLB English-Swahili translation data from FAIR \citep{nllb-2022-no} & \footnotesize \url{https://huggingface.co/datasets/allenai/nllb} \\ \cline{2-3}
     & Inkuba dataset from Lelapa \citep{tonja2024inkubalm} & \footnotesize \url{https://huggingface.co/datasets/lelapa/Inkuba-instruct} \\ \cline{2-3}
     & xP3 MT dataset from BigScience, with \flores samples removed \citep{muennighoff2022crosslingual} & \footnotesize \url{https://huggingface.co/datasets/bigscience/xP3mt} \\ \hline
    
    \end{tabular}
    \end{center}
\end{table}

All datasets listed above were verified to be used in compliance with their respective licenses. Each dataset was properly attributed according to its license requirements.

\subsection{CPT Datasets} \label{cptdata}
\begin{table}[h!]
    \begin{center}
    \caption{Datasets used for continual pretraining (CPT) in this project}
    \begin{tabular}{|c|p{6cm}|p{7cm}|}
    \hline
    \textbf{Category} & \textbf{Datasets} & \textbf{URL} \\ \hline
    
    \multirow{1}{*}{Math} 
     & Open Web mathematical texts collected by the University of Toronto and Cambridge \citep{paster2024openwebmath} & \footnotesize \url{https://huggingface.co/datasets/open-web-math/open-web-math} \\ \hline
    
    \multirow{1}{*}{Bengali} 
     & The ROOTS corpus subset of Bengali Wikipedia from BigScience \citep{laurencon2022the} & \footnotesize \url{https://huggingface.co/datasets/bigscience-data/roots_indic-bn_wikisource} \\ \hline
     
    \end{tabular}
    \end{center}
\end{table}
All datasets listed above were verified to be used in compliance with their respective licenses. Each dataset was properly attributed according to its license requirements. 

\subsection{Off-the-shelf Model Results} \label{shelf}

To motivate the use of our four models and the three target languages, we provide preliminary results of these models prior to any fine-tuning.

\begin{table}[h!]
    \centering
    \begin{tabular}{|cc||c|c|c|c||c|c|c|c|}
        \hline
        \textbf{Model} & \textbf{Size} & \multicolumn{4}{c||}{\textbf{MGSM}} & \multicolumn{4}{c|}{\textbf{\belebele}} \\
         & & \textbf{EN} & \textbf{SW} & \textbf{BN} & \textbf{TE} & \textbf{EN} & \textbf{SW} & \textbf{BN} & \textbf{TE} \\
        \hline \hline
        \llama   & 8B   & 79.6\% & 52.0\% & 32.8\% & 11.2\% & 88.6\% & 56.1\% & 59.3\% & 53.6\% \\ \hline
        \qwen    & 7B   & 76.8\% & 12.8\% & 37.6\% & 13.6\% & 91.1\% & 37.2\% & 64.7\% & 41.3\% \\ \hline
        \aya     & 8B   & 78.8\% & 10.8\% & 21.6\% & 3.2\%  & 81.6\% & 32.3\% & 42.3\% & 29.9\% \\ \hline
        \falcon  & 7B   & 78.4\% & 14.4\% & 10.4\% & 3.6\%  & 85.9\% & 36.3\% & 34.8\% & 30.1\% \\ \hline
    \end{tabular}

    \caption{The results on the MGSM (2-shot, EM accuracy ($\uparrow$)) and \belebele (0-shot accuracy ($\uparrow$)) benchmarks for the four models used in our experiments. We note that for all models, we use the instruction-finetuned version.}
\end{table}



\end{document}